\documentclass[preprint,3p,times]{elsarticle}

\usepackage{amssymb}
\usepackage{amsmath}

\usepackage{multirow}
\usepackage[colorlinks,
            linkcolor=blue,
            anchorcolor=blue,
            citecolor=blue]{hyperref}

\journal{Applied Soft Computing}

\begin{document}

\begin{frontmatter}

\title{Hierarchical Ranking Neural Network for Long Document Readability Assessment}

\author[1]{Yurui Zheng} 
\ead{846006629@qq.com}

\author[1]{Yijun Chen} 
\ead{1051575227@qq.com}

\author[1]{Shaohong Zhang \corref{cor1}}
\cortext[cor1]{Corresponding author}
\ead{zimzsh@qq.com}

\affiliation[1]{organization={School of Computer Science and Cyber Engineering},
                addressline={Guangzhou University}, 
                city={Guangzhou}, 
                postcode={510000}, 
                state={Guangdong},
                country={China}}

\begin{abstract}
Readability assessment aims to evaluate the reading difficulty of a text. In recent years, while deep learning technology has been gradually applied to readability assessment, most approaches fail to consider either the length of the text or the ordinal relationship of readability labels. This paper proposes a bidirectional readability assessment mechanism that captures contextual information to identify regions with rich semantic information in the text, thereby predicting the readability level of individual sentences. These sentence-level labels are then used to assist in predicting the overall readability level of the document. Additionally, a pairwise sorting algorithm is introduced to model the ordinal relationship between readability levels through label subtraction. Experimental results on Chinese and English datasets demonstrate that the proposed model achieves competitive performance and outperforms other baseline models.
\end{abstract}


\end{frontmatter}

\section{Introduction}

Automatic Text Readability (ARA) research originated in the early 20th century, aiming to evaluate text reading difficulty and assist educators in recommending appropriate reading materials for learners ~\cite{vogel1928objective}. Readability assessment approaches are generally classified into three paradigms: human evaluation, co-selection-based analysis, and content-based analysis. Human evaluation involves expert annotation or reader surveys; co-selection methods leverage user interaction data such as reading time or choices~\cite{cop2015eye}; and content-based approaches infer readability using linguistic, syntactic, or semantic features extracted from the text itself. Early studies predominantly relied on experts’ subjective evaluations and simple statistical features, such as sentence length and word complexity. However, these approaches suffered from high subjectivity, with evaluation results often varying depending on the evaluators’ criteria and purposes. The current mainstream evaluation is based on content-based analysis. In the field of natural language processing (NLP), readability assessment research was initially limited. Early researchers developed readability formulas by designing simple linguistic features ~\cite{flesch1948new,dale1948formula,mc1969smog}. As the number of readability formulas grew, the application of readability analysis expanded from education to diverse industries, including law ~\cite{villata2020plain}, medicine ~\cite{sare2020readability,perni2019assessment}, and government policy to improve document intelligibility.

Interest in this issue among NLP researchers has only emerged in the past two decades. From statistical language models and feature-engineering-based machine learning methods to more recent deep neural networks, a range of methods has been explored for this task~\cite{vajjala2021trends}. Existing works~\cite{schwarm2005reading,lee2021pushing,hansen2021machine,pilan2016predicting} primarily involve designing diverse language features and employing machine learning models to text classification, such as support vector machines(SVM) or multilayer perceptron.

In recent years, the proliferation of deep neural networks has advanced text readability research, circumventing the need for cumbersome linguistic feature engineering. However, deep neural networks have not outperformed traditional machine learning models by a wide margin. One key challenge is the highly variable text lengths in readability evaluation datasets ~\cite{li2022unified}—upper-grade texts are typically longer, while lower-grade texts are shorter, a discrepancy that is particularly pronounced in Chinese corpora. Most current models ~\cite{martinc2021supervised,imperial2021bert,lee2023prompt,risch2020bagging} employ pre-trained BERT ~\cite{kenton2019bert} as word encoders, but BERT’s maximum input length is limited to 512 tokens. To address this, the hierarchical attention network (HAN) framework ~\cite{yang2016hierarchical} offers a solution: its word-to-sentence-to-document hierarchical structure enables models to capture long-text dependencies more effectively.

Additionally, while the text readability task is commonly framed as a text classification problem, ~\citet{zheng2019hybrid} highlights the critical role of context vectors in traditional attention mechanisms. However, prior studies ~\cite{martinc2021supervised} either neglect context vectors or initialize them randomly, a practice that undermines contextual modeling. To address this, we introduce a multi-dimensional context weight vector, which generates sentence-level representation vectors by aggregating context weights and word embeddings. Notably, each text is assigned a unique context weight vector to capture its global semantic characteristics.

Moreover, this paper innovatively proposes a bidirectional text readability assessment framework based on hierarchical modeling. The proposed method first employs document-level readability annotations to infer sentence-level readability labels via a hierarchical model, thereby constructing a sentence-level readability corpus. Subsequently, this sentence corpus is used as auxiliary supervision to enhance the model’s forward prediction of document-level readability. This approach not only enables the modeling of implicit sentence-level readability features, but also enhances the model’s contextual understanding by leveraging hierarchical information, leading to a more fine-grained readability assessment system.

Due to the orderliness brought by the text difficulty level labels, there are also some works that treat it as a regression task~\cite{azpiazu2019multiattentive} or a ranking task~\cite{lee2022neural}. Because texts in adjacent grades tend to be more similar than distant grades, but do not distinguish them in the classification task. Therefore, we construct the Ranking Model in the tail of the model, which adopts the way of pairwise comparison and label subtraction to learn the order relationship between categories.

Our research contributions can thus be concluded as follows:
\begin{itemize}
\item Compared with English, research on Chinese text readability remains relatively limited. We constructed a Chinese corpus based on textbooks from mainland China and developed a feature engineering scheme tailored to the Chinese language domain.

\item We designed a hierarchical model inspired by hierarchical neural architectures, which preserves more information from long documents. Meanwhile, we introduced multi-dimensional contextual weighting to guide the attention mechanism in identifying informative words within the input sequence.

\item We proposed a bidirectional readability assessment framework, which utilizes sentence-level readability labels to further enhance the model's ability to predict text-level readability in the forward direction.

\item At the end of the forward prediction module, we introduced a ranking model that learns the ordinal relations between readability levels by modeling label differences, and outputs the optimal text ranking through pairwise comparisons.
\end{itemize}

The rest of this paper is structured as follows:
\begin{itemize}
\item Section \ref{setc2} provides a review of related work.
\item Section \ref{setc3} introduces our proposed model.
\item Sections \ref{setc4} and \ref{setc5} describe the datasets and experimental setup, respectively.
\item Section \ref{setc6} discusses the experimental results.
\item Section \ref{setc7} concludes the paper and outlines future directions.
\end{itemize}

\section{Related Work \label{setc2}}

Early research focused on developing the readability formula, which is a weighted linear function, including Dale-Chall~\cite{dale1948formula}, SMOG~\cite{mc1969smog}, and Flesch-Kincaid~\cite{kincaid1975derivation}. Using readability formula to evaluate text difficulty is objective and easy to calculate, but it only considers simple text features, and does not consider language features well. Nonetheless, the traditional readability formulation also laid the foundation for later readability research.

In machine learning, researchers train ARA models with the help of classifiers by constructing a large number of linguistic features. ~\citet{heilman2008analysis} used a combination of lexical features and grammatical features that are derived from subtrees of syntactic parses, while also verifying that ordinal regression models were most effective in predicting reading difficulty. ~\citet{feng2010comparison} employed SVM and logistic regression to compare and evaluate several sets of explanatory variables - including shallow, language modeling, POS, syntactic, and discourse features, and checked that the judicious combination of various features led to significant improvements over the state of the art. ~\citet{hancke2012readability} developed new morphological features and achieved 89.7\% accuracy in German readability classification based on these features. ~\citet{qiu2018exploring} designed 100 factors to systematically evaluate the influence of four levels of linguistic features (shallow, part of speech, syntax and discourse) on the difficulty of predicting texts for L1 Chinese learners, and further selected 22 features with regression significance. ~\citet{deutsch2020linguistic} and ~\citet{lee2021pushing} similarly leverage various language-driven features combined with simple machine learning models and aided by deep learning models to improve performance.

Deep learning methods are becoming more and more widely used in ARA. ~\citet{jiang2015graph} proposed a graph-based readability evaluation method using word coupling, which combines the merits of word frequencies and text features. ~\citet{azpiazu2019multiattentive} present a multiattentive recurrent neural network architecture for automatic multilingual readability assessment. This architecture considers raw words as its main input, but internally captures text structure and informs its word attention process using other syntax- and morphology-related datapoints, known to be of great importance to readability.  ~\citet{blaneck2022automatic} studied the ability of ensembles of fine-tuned GBERT and GPT-2-Wechsel models to reliably predict the readability of German sentences.

ARA can similarly be viewed as a regression or ordinal regression task due to the orderliness of the readability labels. ~\citet{meng2020readnet} proposed a new comprehensive framework that uses a hierarchical self-attention model to analyze document readability. In this model, the goal is to minimize the ordinal regression loss ~\cite{rennie2005loss}. ~\citet{lee2022neural} proposed the first neural pairwise ranking model for ARA and showed the first results of cross-lingual, zero-shot evaluation of ARA using neural models. Z~\citet{zeng2022enhancing} used soft labels ~\cite{diaz2019soft} to exploit the ordinal nature of the readability assessment task.In addition to the above label modeling-based methods, ~\citet{tanaka2010sorting}  proposed a readability assessment framework based on a sorting mechanism. Instead of directly predicting the absolute readability level, a binary comparator is trained to judge the relative readability between any two texts, and then the text set is sorted by a sorting algorithm. The biggest advantage of this method is that the training data only needs two levels (easy and difficult), which greatly reduces the difficulty of annotation. It is particularly suitable for low-resource language environments where training data is scarce and level annotation is difficult.

In experiments across corpora, ~\citet{xia2019text} applied a generalization method to adapt models trained on larger native corpora to estimate text readability for learners, and explored domain adaptation and self-learning techniques to make use of the native data to improve system performance on the limited L2 data. ~\citet{madrazo2020cross} developed a cross-lingual readability assessment strategy that serves as a means to empirically explore the potential advantages of employing a single strategy (and set of features) for readability assessment in different languages, including interlanguage prediction agreement and prediction accuracy improvement for low-resource languages.

\section{Methodology \label{setc3}}

\begin{figure*}[!t]
\centering
\includegraphics[width=0.6\textwidth]{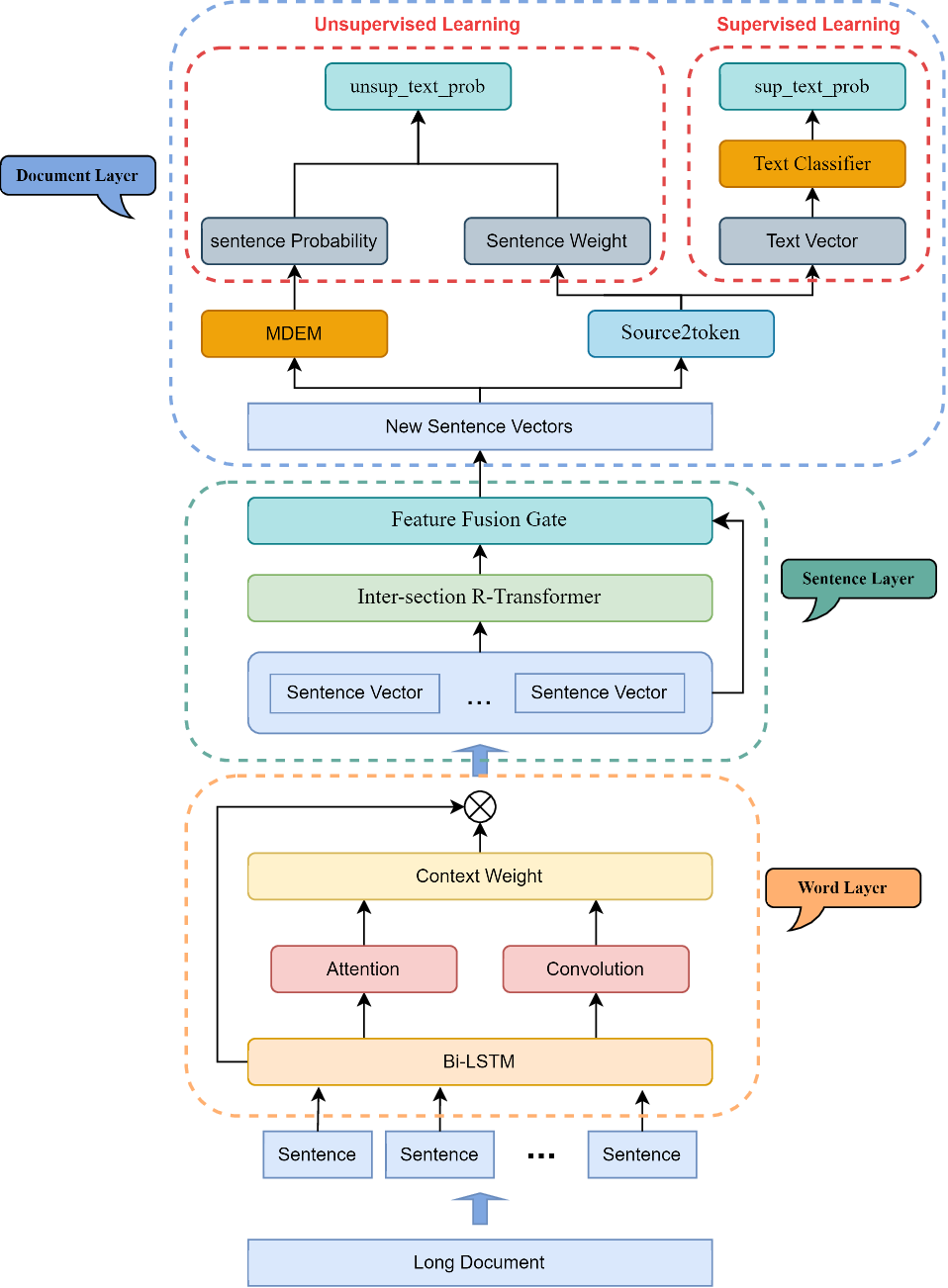}
\caption{The overall structure of the HHNN-MDEM.}
\label{fig:1}
\end{figure*}

First, we built a set of linguistic features for the Chinese corpus. Then, we proposed an ARA hierarchical model that can be used to evaluate the readability level of sentences in long documents, and introduced a multidimensional context weight vector and a multi-head difficulty embedding matrix(MDEM) into the model, aiming to solve the information loss problem in the traditional attention mechanism and how to reversely predict the readability level of sentences through text. We call this model HHNN-MDEM (Hierarchical Hybrid Neural Network with Multi-Head Embedding Matrix). The overall structure of the model is shown in Fig.~\ref{fig:1}.

\subsection{Explicit Features}

Compared with English, there are few researches on Chinese readability feature engineering. Therefore, we construct Chinese traditional features, including lexical features, part-of-speech features, discourse features and article cohesion features, with reference to relevant traditional features ~\cite{qiu2018exploring,sung2013investigating,ma2022research}. In the specific domain of classical Chinese, we construct thematic features and complex semantic features for this purpose. See the appendix for a detailed explanation.

For traditional features in English, explicit feature extraction comes from lingfeat~\cite{lee2021pushing}, which studies 255 language features. For already existing features, variants were added to expand coverage, including the development of high-level semantic features related to the topic: semantic richness, clarity, and noise. In order to improve the generalization of the experiment, the features related to the dataset were removed in this experiment, that is, the topic knowledge features of WeeBit and OneStopEnglish, a total of 32 features. Since there are 16 Entity Grid Features missing from some of the features generated by lingfeat, 207 explicit features were finally generated.

\subsection{Proposed Model}

As shown in Figure.~\ref{fig:1}, HHNN-MDEM is a hierarchical semi-supervised learning framework that uses a hybrid supervision and hierarchical consistency training strategy to achieve automatic labeling of sentence readability labels. The hierarchical learning architecture can be divided into three components, namely the word layer, sentence layer, and document layer.

\subsubsection{Word Layer}

In order for the model to learn more text information from long documents, we divide the text into multiple sentences. Specifically, for a given long document, we represent it as a sequence of sentences $S=\{s_{1},\ldots,s_{n} \}$, where $s_{i}=\{w_{i,1},\ldots,w_{i,m}\}$, $i=1,\ldots,n$ denotes the token sequence of the $i^{th}$ sentence. If necessary, each sentence will be padded and cut to maintain the same length. The Embedding layer encodes each word $w_{i,j}$ into a d-dimensional vector based on word embeddings. The output is an $m\times d$ dimensional matrix $A^{i} = [\mathbf{e}_{1}, \ldots, \mathbf{e}_{m}]$, where $d$ is the embedding dimension and $m$ is the number of words in the $i^{th}$ sentence.

Due to the orderliness of text sequences, we use a bidirectional recurrent neural network, such as LSTM or GRU, to capture the order information between words. Specifically, the word vectors of $A^{i}$ are sequentially input into the Bi-LSTM. By connecting the forward hidden state and the backward hidden state, a new word-level representation $h_i^t$ is produced, denoted as $\mathbf{h}_{i}^{t} = [h_{i,1}^{t}, \ldots, h_{i,m}^{t}] \in \mathbb{R}^{m \times d}$, i.e., $\mathbf{h}_{i}^{t} = Bi-LSTM(A^{i})$.

\begin{figure}[!t]
\centering
\includegraphics[scale=.7]{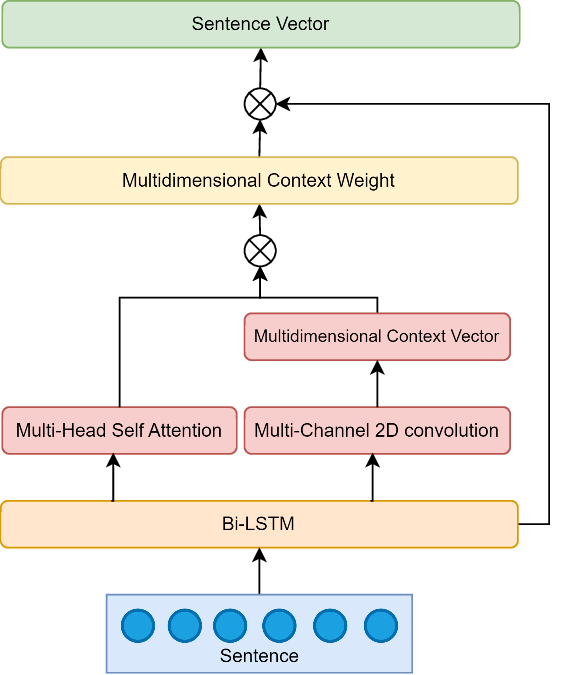}
\caption{Overall structure of word layer.}
\label{fig:2}
\end{figure}

{\bf Multidimensional context weights.} The multi-head self-attention mechanism is used to learn the interaction between words as the attention coefficient. Then, a multi-channel two-dimensional convolutional network is used to learn the multi-dimensional context vector of the sentence, and the attention coefficient and the multi-dimensional context vector are multiplied as the final multi-dimensional context weight of the word. The representation vector of each sentence is formed by aggregating the multi-dimensional context weights and the words. The overall structure is shown in Fig.~\ref{fig:2}.

~\citet{du2019convolution} proved that the convolution of CNN is precisely the process of calculating the similarity between text and the \textit{attentivesearch templates}. Therefore, the CNN layer extracts the most influential n-gram syntax of different semantic aspects from the text and uses them as context vectors. According to the output of the Bi-LSTM, the local features on $h_i^t$ are extracted using convolutional kernels. The convolution operation involves a set of $k$ convolutional kernels, where each convolutional kernel $Conv \in \mathbb{R}^{l \times d}$ is applied to a window of $l$ words, generating new features $c_j$ from the window of the vector $h_{i,j:j+l-1}^t$, with $j = 1, \ldots, m$ representing the $j^{th}$ word of the $i^{th}$ sentence, as follows,

\begin{equation}
    c_j=Conv \cdot h_{i,j:j+l-1}^t+b
\end{equation}
where is $b$ bias. This convolution kernel is applied to each possible window of the matrix $h^t$ to generate the context vector $\mathbf{\hat{c}}=[c_1,\ldots,c_{m-l+1}]$. By employing k convolutional kernels, the context vectors are concatenated together, resulting in a k-dimensional context vector $C = [\mathbf{\hat{c}}_1, \ldots, \mathbf{\hat{c}}_k] \in \mathbb{R}^{(m-l+1) \times k}$.

Multi-head attention (MHA)~\cite{vaswani2017attention} allows the model to jointly attend to information from different representational subspaces at different positions. The multi-head self-attention mechanism (MHSA) is a special case of MHA, where the queries ($Q$), keys ($K$), and values ($V$) of the self-attention layer all come from the output of the previous encoder layer, i.e., the inputs are $Q=K=V$. Here, we modify this mechanism by taking $Q$ and $K$ from the output $h^t$, and by calculating the similarity between $Q$ and $K$, we learn the interactions between words to obtain the attention coefficients. The values $V$, on the other hand, are taken from the multidimensional contextual vector $C$. Thus, the multidimensional contextual weights $W_i$ are calculated using the weighted sum of the attention coefficients and the multidimensional contextual vector $C$. The specific calculation is shown as follows,

\begin{equation}
    W_i^t=softmax(ReLU(MHSA(\mathbf{h}_i^t,\mathbf{h}_i^t,C)))
\end{equation}
where $softmax$ refers to the along-column normalization and $ReLU$ is the modified linear unit activation function. Finally, by taking the weighted sum of $h_i^t$ and $W_i^t$, the sentence representation vector $h_i^s$ is obtained, as represented below,

\begin{equation}
    \mathbf{h_i^s}=\sum_{j=1}^{m} W_{i,j}^t \odot h_{i,j}^t
\end{equation}
where $\odot$ represents the element wise product of two matrices. We denote $h^s = [\mathbf{h}_1^s, \ldots, \mathbf{h}_n^s] \in \mathbb{R}^{n \times d}$. According to the multi-dimensional context vector generated by the convolutional network, the interactive word vector of multi-head self-attention is combined with the context vector to generate multi-dimensional context weights. The weights select useful local features from the recurrent layer. This hybrid network retains the advantages of the three models, and each text has its own context vector.

\subsubsection{Sentence Layer}

In order to further capture the long-term dependency of different sentences, we adopt the Inter-section R-Transformer~\cite{hu2021hierarchical}, which adopts N layer of transformers and replaces the residual blocks with residual fusion gates. The multi-head self-attention layer MHSA and normalization are firstly applied to the sentence level features as follows,

\begin{equation}
    o^s=norm(MHSA(h^s))
\end{equation}
where $o^s=[o_1^s,\ldots ,o_n^s]\in \mathbb{R}^{n \times d}$, and $norm$ represents the normalization operation.

To combine the local and global context features at the sentence level, the Inter-section R-Transformer uses a residual fusion gate to dynamically merge the input and output of multi-head self-attention. The output sequence of the residual fusion gate $e^s=e_1^s,\ldots ,e_n^s$ is computed as follows,

\begin{equation}
    G1=sigmoid(W_{11}^s o^s+W_{12}^s h^s+b_1^s )
\end{equation}

\begin{equation}
    e^s=G1 \odot h^s+(1-G1) \odot o^s
\end{equation}
where $W_{11}^s$, $W_{12}^s$ and $b_1^s$ are the parameters of the first residual fusion gate and $sigmoid$ is the activation function. The output of gate $e^s$ is then passed through a fully connected layer $f^s$ followed with a normalization process. The final output of the Inter-section R-Transformer is obtained by applying another residual fusion gate, which can be expressed as,

\begin{equation}
    G2=sigmoid(W_{21}^s norm(f^s (e^s ))+W_{22}^s e^s+b_2^s )
\end{equation}

\begin{equation}
    v^s=G2\odot e^s+(1-G2)\odot norm(f^s (e^s ))
\end{equation}
where $W^{21}$, $W^{22}$, and $b_2^s$ are the parameters of the second residual fusion gate.

We employ feature fusion gate~\cite{hu2021hierarchical} to fuse $h^s$ and $v^s$ to generate the final feature representation of the sentence. The formula of the feature fusion gate is as follows,

\begin{equation}
    F=ReLU(W_3^s [h^s,v^s ]+b_3^s )
\end{equation}

\begin{equation}
    G=sigmoid(W_4^s [h^s,v^s ]+b_4^s )
\end{equation}

\begin{equation}
    u^s=G \odot F+(1-G)\odot h^s
\end{equation}
where $W_3^s \in \mathbb{R}^{2d \times d}$, $W_4^s \in \mathbb{R}^{2d \times d}$, $b_3^s \in \mathbb{R}^d$, $b_4^s \in \mathbb{R}^d$ are the learning parameters of the feature fusion gate. $u^s=[u_1^s, \ldots ,u_n^s] \in \mathbb{R}^{n \times d}$ is the output of the feature fusion gate.

\subsubsection{Document Layer}

We apply the sentence vector $u^s$ to the "source2token" self-attention~\cite{shen2018disan}, explore the dependencies between sentences, and compress $u^s$ into a document vector $d$, which is represented as follows,

\begin{equation}
    W^D=softmax(W_1^d ReLU(W_2^d u^s+b_1^d )+b_2^d )
\end{equation}

\begin{equation}
    d=\sum_{i=1}^n W_i^D \odot u_i^s
\end{equation}
where $W_1^d \in \mathbb{R}^{d \times d}$, $W_2^d \in \mathbb{R}^{d \times d}$, $b_1^d \in \mathbb{R}^d$, $b_2^d \in \mathbb{R}^d$ are the learning parameters of the "source2token" self-attention module.

{\bf Supervised Learning.} The objective of the supervised component is to leverage document-level labels as explicit supervision signals to train a text classifier that learns document-level feature representations. This enables the effective transfer of document-level label information to the sentence level, driving the model to learn semantic mappings from documents to sentences and laying a solid foundation for the subsequent automatic generation of sentence-level labels. Specifically, the input text is first encoded by the HHNN model to obtain a document representation $d$. Then, $d$ is passed through a fully connected layer to map it to a probability distribution $\mathbf{r}$ over readability levels. The supervised training objective is to minimize the cross-entropy loss, as defined by the following formula,

\begin{equation}
\mathcal{L}_{\text {sup }}=-\frac{1}{N} \sum_{i=1}^{N} \sum_{k=1}^{Y} y_{i k} \log \left(r_{i k}\right)
\end{equation}
where $N$ denotes the number of samples, $y_{i k}$ represents the ground-truth label of sample $i$ for class $k$, and $r_{i k}$ denotes the corresponding predicted probability. The supervised component not only trains a document-level classifier but also ensures that the model, optimized through cross-entropy loss, captures explicit semantic information from the document. This helps align the sentence representation space with the semantic structure of readability levels.

{\bf Unsupervised Learning.} The core objective of the unsupervised learning component is to learn sentence-level readability labels through hierarchical propagation of document-level labels. To achieve this, we introduce the Multi-Head Difficulty Embedding Matrix (MDEM) module, which, combined with hierarchical consistency training, enables the automatic generation of sentence-level labels.

The MDEM module learns the relative difficulty scores of sentence representations with respect to different readability levels. At its core, it employs a multi-head attention mechanism to compute the difficulty scores of each sentence under each readability category, thereby producing a difficulty distribution for each sentence. Specifically, the MDEM module performs a matrix multiplication between the sentence representations $u^{s} \in \mathbb{R}^{n \times d}$ and the multi-head difficulty embedding matrix $M \in \mathbb{R}^{h \times z \times Y}$ , where each head learns category-specific difficulty features. The sentence vectors $u^{s}$ are first projected into multi-head format, as formulated below,

\begin{equation}
u^{s^{\prime}}=\operatorname{reshape}\left(u^{s}\right) \in \mathbb{R}^{h \times n \times z}
\end{equation}
here, $h$ denotes the number of attention heads, $n$ is the number of sentences, and $z$ represents the dimension of each input feature, where $z=d/h$.
Next, the sentence representations are multiplied with MDEM to obtain the sentence-level scores under each readability category across different heads $a=\left[a_{1}, \ldots, a_{h}\right] \in \mathbb{R}^{h \times n \times Y}$.Finally, a summation is performed across the multi-head dimension to obtain the aggregated sentence-level score matrix $A$,

\begin{equation}
A=\sum_{i=1}^{h} a_{i}
\end{equation}
$A$ reflects the difficulty scores of each sentence across different readability categories.

The prediction result of the sentence label is not provided directly, but the information of the document-level label is propagated to the sentence level through the reverse label, and finally obtained through the MDEM module. In order to ensure the accuracy and consistency of the sentence label, the experiment introduced the KL divergence to measure the probability consistency between the sentence label and the document label. First, the sentence score matrix A is weighted summed to obtain the text score vector $\mathbf{\hat{r}}$, and the weight comes from the "source2token" self-attention mechanism at the document word level. Then, the text score vector $\mathbf{\hat{r}}$ is processed by LogSoftmax and compared with the document prediction probability $\mathbf{d}$ obtained in supervised learning. The probability distribution of the two is constrained by the KL divergence. The specific formula is as follows,

\begin{equation}
\mathcal{L}_{\text {unsup }}=\frac{1}{N} \sum_{i=1}^{N} D_{K L}(\operatorname{LogSoftmax}(\mathbf{\hat{r}}) \| \mathbf{d})
\end{equation}

KL divergence is used as the loss function of the unsupervised part to directly correct the weights of the MDEM module and the sentence encoder, improve the distribution alignment capability, and ensure the semantic consistency between sentence-level labels and document labels. It helps the model learn semantic features consistent with document labels from unlabeled sentences and further optimizes the prediction accuracy of sentence labels. It also forms a two-way closed loop of "document label-driven sentence annotation—sentence feature reconstruction document prediction" with supervised learning.

{\bf Hierarchical consistency training.} A hybrid supervision strategy is adopted during the training process. When supervised learning and unsupervised learning are used for joint training, a weighting factor $\lambda$ is used to balance the supervised cross entropy and unsupervised consistency training losses. Formally speaking, the complete training objective formula is as follows,

\begin{equation}
\mathcal{L}=\mathcal{L}_{\text {sup }}+\lambda \mathcal{L}_{\text {unsup }}
\end{equation}

In the hierarchical consistency training, the experiment adopted three training techniques proposed by Google UDA~\cite{xie2020unsupervised} to ensure that the model establishes a stable and consistent semantic mapping between the document level and the sentence level. These techniques are Training Signal Annealing (TSA), Confidence-based Masking, and Sharpening Predictions.

\subsection{Forward Text Readability Assessment}

\begin{figure}[!t]
\centering
\includegraphics[scale=.7]{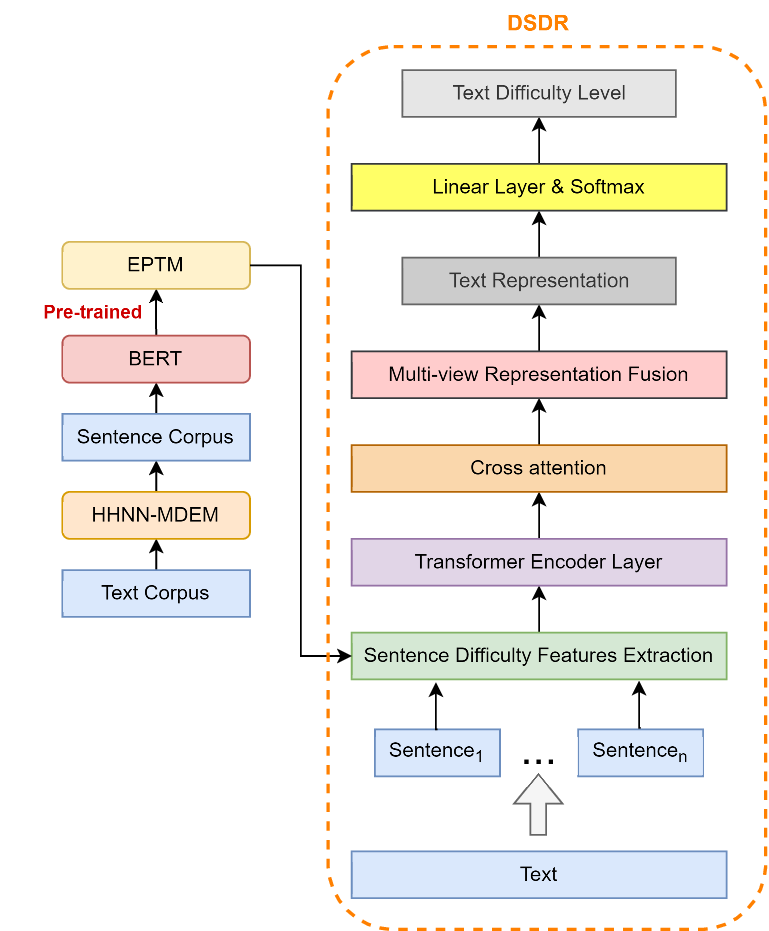}
\caption{The overall structure of the DSDR.}
\label{fig:3}
\end{figure}

After using the sentence corpus obtained by HHNN-MEDM, sentence labels are used to assist in predicting the document-level readability level. Specifically, the structure of the DSDR~\cite{1023569814.nh} model is borrowed, as shown in the Fig.~\ref{fig:3}, and combined with the Ranking Model, DSDRRM is proposed.

\subsubsection{DSDR}

First, the pre-trained BERT model is directly used to train the sentence corpus to obtain an enhanced pre-trained model, which combines the difficulty multi-view representation and multi-view representation fusion, and can effectively apply sentence labels to document-level readability prediction, thereby further improving the accuracy of readability evaluation.

{\bf Sentence-level difficulty-aware pre-training.} Use BERT to perform supervised training on a sentence corpus and build an enhanced pre-training model (EPTM) to learn the difficulty representation capability of sentences.

{\bf Difficulty multi-view representation.} EPTM is used to generate the representation of each sentence, and the context information is supplemented by the Transformer encoder to form a sentence-level context representation. Furthermore, multiple learnable difficulty vectors $C\in\mathbb{R}^{m\times d}$ are introduced, where m is the number of difficulty categories and d is the dimension of the sentence vector. The semantic representation of sentences at different difficulty levels is extracted through the cross-attention mechanism to form a multi-view representation $R=Attention\left(CW^Q,H^tW^k,H^tW^V\right)$.

{\bf Multi-view representation fusion.} The multi-view difficulty representations are fused by average pooling to obtain the document-level representation $T$. 

\subsubsection{Ranking Model}

In ARA prediction tasks, the order relationship between categories is important, and there is an obvious order relationship between adjacent difficulty labels, which is difficult to capture by simple classification tasks, because classification tasks usually assume that categories are equally weighted and independent. Therefore, we propose the Ranking Model, which typically requires the model to output a specific category of the corresponding sample for classification tasks. The Ranking Model further attempts to predict the order relationship of categories.

\begin{figure}[!t]
\centering
\includegraphics[scale=.5]{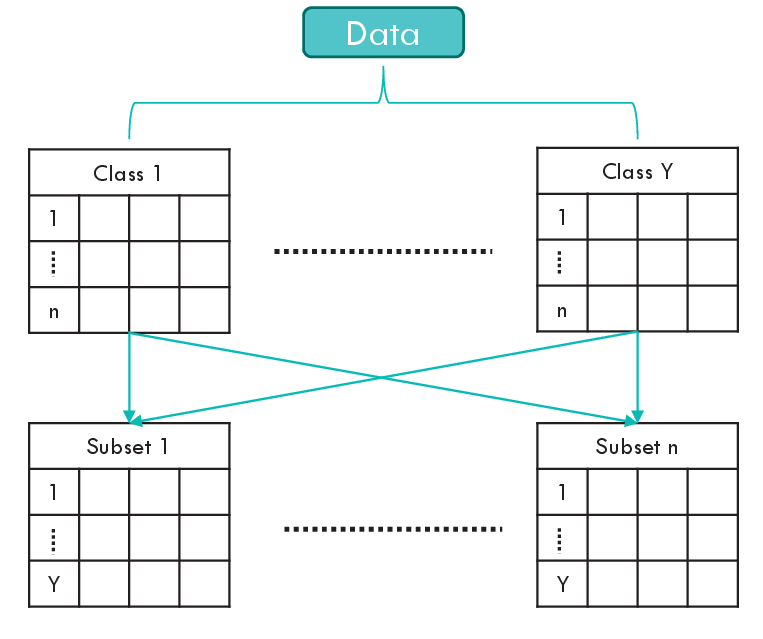}
\caption{Data subset construction process. Suppose we have readability rank Y ordered categories in our dataset, each with n samples.}
\label{fig:4}
\end{figure}

Firstly, based on the original data set, a number of data subsets are constructed according to the process in Fig.~\ref{fig:4}. Each data subset contains samples corresponding to the number of readability levels. That is, there are as many samples in the data subset as there are classes in the readability level, and there are no samples of the same level in the subset. Then, the samples in each subset are combined by pairwise permutation, and the difference in difficulty labels of the samples is used as the new label. This comparison allows the model to learn the differences between neighboring categories and thus better understand the sequential relationships between categories. In addition, this pairwise comparison method can also greatly expand the amount of data, that is, each data subset can construct $B_Y^{Y-1}=Y \times (Y-1)$ samples. The combined two samples are concatenated $d_{concat}=[d_1;d_2]$ into a fully connected layer to act as a classifier. Finally, each data subset was put into the model as a batch to continue training, and cross-entropy loss was used as the training target. By randomly combining samples with samples from different data subsets, a single sample will have multiple prediction results, which increases the error tolerance of each sample. Finally, the final readability level of each sample is obtained by hard voting.

As for the test set, we combined the data of the test set with the data subset of the training set to form a pairwise comparison. By predicting the label difference and adding it to the sample label in the data subset, we finally get the readability grade of the sample in the test set. Similarly, each sample grade in the test set is obtained by hard voting.

\section{Data \label{setc4}}

In order to verify the effectiveness of our proposed model, we conduct experiments on five datasets of long documents, including two English datasets and three Chinese datasets. We split the training and test sets with an 8:2 ratio across all datasets. The dataset statistical distribution is shown in Table~\ref{tab1} and Table~\ref{tab2}.

\begin{table}[!t]
\caption{\label{tab1}Statistics of datasets for OSP,CEE, and CTRDG. Avg.Length is the average number of words per text.}
\centering
\begin{tabular}{ccccccc}
\hline
\multirow{2}{*}{\textbf{Grade}} & \multicolumn{2}{c}{\textbf{OSP}} & \multicolumn{2}{c}{\textbf{CEE}} & \multicolumn{2}{c}{\textbf{CTRDG}} \\ \cline{2-7} 
                                & Texts        & Avg.Length        & Texts        & Avg.Length        & Texts         & Avg.Length        \\ \hline
\textbf{1}                      & 189          & 519               & 64           & 140               & 714           & 14.12                \\
\textbf{2}                      & 189          & 654               & 60           & 268               & 1102           & 23.32               \\
\textbf{3}                      & 189          & 809               & 71           & 613               & 1310           & 40.69               \\
\textbf{4}                      & -            & -                 & 67           & 767               & 971             & 85.26                 \\
\textbf{5}                      & -            & -                 & 69           & 752               & 1163             & 235.28                 \\ 
\textbf{6}                      & -            & -                 & -           & -                  & 461             & 580.57                 \\ \hline
\end{tabular}
\end{table}

\begin{table}[!t]
\caption{\label{tab2}Statistics of datasets for CMER, and CLT. Avg.Length is the average number of words per text.}
\centering
\begin{tabular}{ccccc}
\hline
                                 & \multicolumn{2}{c}{\textbf{CMER}} & \multicolumn{2}{c}{\textbf{CLT}}         \\ \cline{2-5} 
\multirow{-2}{*}{\textbf{Grade}} & Texts         & Avg.Length        & Texts                       & Avg.Length \\ \hline
\textbf{1}                       & 218           & 164               & 107                         & 112        \\
\textbf{2}                       & 217           & 347               & 181                         & 191        \\
\textbf{3}                       & 234           & 604               & 203                         & 308        \\
\textbf{4}                       & 229           & 699               & 192                         & 429        \\
\textbf{5}                       & 199           & 757               & 171                         & 534        \\
\textbf{6}                       & 255           & 775               & 155                         & 651        \\
\textbf{7}                       & 221           & 1352              & 91                          & 1237       \\
\textbf{8}                       & 204           & 1409              & 85                          & 1124       \\
\textbf{9}                       & 187           & 1429              & 61                          & 1927       \\
\textbf{10}                      & 100           & 2384              & -                           & -          \\
\textbf{11}                      & 95            & 2418              & -                           & -          \\
\textbf{12}                      & 97            & 2226              & -                           & -          \\ \hline
\end{tabular}
\end{table}

{\bf OneStopEnglish(OSP)}~\cite{vajjala2018onestopenglish} is a parallel corpus that can be used for automatic readability assessment and automatic text simplification. The corpus consists of 189 texts, each with three versions (567 texts in total).

{\bf Cambridge(CEE)}~\cite{xia2019text} is a corpus for L2 learners that contains reading articles from five major Cambridge English examinations (KET, PET, FCE, CAE, CPE). The five exams are aimed at learners at levels A2-C2 of the Common European Frame of Reference.

{\bf CMER}~\cite{zeng2022enhancing} consists of texts from extracurricular reading books for kids and teenagers at China mainland currently on the book market, with a total of 3,395,923 characters, distributed in 2,260 texts in 12 levels.

{\bf CLT} is a dataset of Chinese language textbooks that we collected from primary and secondary school textbooks from multiple publishing houses. All the Chinese textbooks are taken from the first grade of primary school to the third grade of junior high school, with a total of 9 grades.

{\bf CTRDG}~\cite{tan2024multi} is a dataset about the Chinese Proficiency Test (HSK), with a total of 6 levels.

\section{Experimental Setup \label{setc5}}

This section presents the comparison baselines with our proposed model, the evaluation metrics, and the detailed details of the experimental implementation.

\subsection{Statistical classification algorithms}

This baseline is based on the explicit features of traditional classifiers including {\bf Logistic Regression (LR)}, {\bf Random Forest (RF)}, and {\bf Support Vector Machines (SVM)}. In Section 3.1 we introduce the traditional features about Chinese and English. The model was implemented using the scikit-learn~\cite{pedregosa2011scikit} tool, and the hyperparameters were dominated by default Settings.

\subsection{Neural document classifiers}

Such baselines represent another line of previous works that employ variants of neural document models for sentence or document classification.

{\bf Vec2Read}~\cite{azpiazu2019multiattentive} uses static word embeddings, Bi-LSTM, word level and sentence level attention mechanisms. Word Attention version of Vec2Read model is adopted in this experiment. The embedding size and hidden layer size of Bi-LSTM are set to 300 and 128, respectively.

{\bf ReadNet}~\cite{meng2020readnet} proposed a new synthesis framework based on transformers that uses a hierarchical self-attention model to analyze document readability. The version of ReadNet model without explicit features is used in this experiment. For article coding, following the settings of the original paper, the number of sentences in each article and the number of words in each sentence were both limited to a maximum of 50, and the number of encoder layers p and q were set to 6. The embedding dimension is d = 100.

{\bf HAN}~\cite{yang2016hierarchical} uses two GRUs, word level and sentence level attention mechanisms to encode word and sentence representations. This experiment uses the same experimental setup as ~\citet{martinc2021supervised}, where the context vector is randomly initialized and the word and sentence embedding sizes are 200 and 100, respectively.

{\bf BERT}~\cite{kenton2019bert} is fine-tuned on English and Chinese using bert-base-uncased\footnote{\url{https://huggingface.co/google-bert/bert-base-uncased}} and bert-base-chinese\footnote{\url{https://huggingface.co/google-bert/bert-base-chinese}} respectively, with the default learning rate of 2e-5.

{\bf DTRA}~\cite{zeng2022enhancing} uses a hierarchical attention network composed of BERT and Bi-LSTM, combined with word level and sentence level attention mechanisms, and performs model pre-training through soft labels of ordinal regression and predicting pairwise relative text difficulty. Since the paper does not provide the corresponding code and experiment detailed details, the results of the paper are extracted directly.

{\bf Lite-DTRA}~\cite{zeng2022enhancing} is a streamlined version of the DTRA model, proposed to reduce the requirements for hardware storage memory. In this version, the pre-trained BERT with frozen parameters is replaced by ALBERT~\cite{lan2019albert}, allowing the model to be trained in an end-to-end manner. Similarly, due to the lack of corresponding code and detailed experimental information provided in the paper, the results of the paper are extracted directly.

\subsection{Training and Evaluation Details}

We used the Pytorch~\cite{paszke2019pytorch} framework for our experiments. In the Embedding layer, the output dimension of the other datasets is 400 except for CMER, which has an output dimension of 512. Similarly, the hidden layer of Bi-LSTM, the number of convolutional kernels of CNN and the number of heads of multi-head self-attention h size in CMER are 256, 256, 16 respectively. 200, 200, 8 in the other datasets, respectively. The window size of the convolution kernel is 3. During training, the learning rate is set to 1e-3, Adam~\cite{kingma2014adam} is used as the optimizer, and the weights decay to 5e-4. TSA was performed in linear form, the $\beta$ threshold based on confidence masking was set to 0.45, the temperature parameter $\tau$ was set to 0.85, the training rounds were set to 30, and the experimental hyperparameters of the DSDR-MDEM model followed the settings of the DSDR model. For evaluation, we calculated precision (acc), adjacent accuracy (adj-acc), weighted F1 score (F1), precision (p), recall (r), and quadratically weighted kappa (qwk). We repeat each experiment three times and report the average score.

Following previous work on readability evaluation, we use qwk as a primary metric to reflect the ordinal alignment between predicted readability levels and ground-truth labels. Although rank correlation metrics such as Spearman's $\rho$ and Kendall's $\tau$ have been recommended for ordinal tasks~\cite{ehara2021evaluation}, qwk provides a widely accepted, label-sensitive alternative that penalizes larger rank discrepancies more severely.

\begin{table*}[!t]
\caption{\label{tab3}Experimental results of readability evaluation on English and Chinese datasets.}
\centering
\scalebox{0.8}{
\begin{tabular}{ccccccccccccc}
\cline{1-12}
Dataset                & Metrics & LR    & RF             & SVM   & VecRead & ReadNet        & HAN    & Bert           & DTRA   & Lite-DTRA & DSDRRM          &  \\ \cline{1-12}
\multirow{6}{*}{OSP}   & acc     & 48.25 & 81.58          & 57.89 & 55.85   & 84.21          & 80.70  & 78.66          & 85.00  & 86.67     & \textbf{89.47}  &  \\
                       & adj-acc & 89.47 & 99.12          & 93.86 & 100.00  & 100.00         & 100.00 & 97.96          & 100.00 & 100.00    & \textbf{100.00} &  \\
                       & F1      & 47.65 & 81.47          & 57.39 & 56.16   & 84.36          & 79.93  & 78.04          & 84.91  & 86.79     & \textbf{89.38}  &  \\
                       & p       & 47.71 & 81.74          & 58.23 & 61.02   & 85.15          & 81.41  & 79.95          & -      & -         & \textbf{89.38}  &  \\
                       & r       & 48.25 & 81.58          & 57.89 & 55.85   & 84.21          & 80.70  & 78.66          & -      & -         & \textbf{89.47}  &  \\
                       & qwk     & 40.05 & 83.81          & 54.56 & 64.60   & 87.50          & 86.19  & 79.59          & -      & -         & \textbf{92.00}  &  \\ \cline{1-12}
\multirow{6}{*}{CEE}   & acc     & 47.76 & 80.60          & 50.75 & 53.03   & 72.73          & 73.74  & 62.63          & -      & -         & \textbf{83.58}  &  \\
                       & adj-acc & 77.61 & 94.03          & 76.12 & 88.38   & 95.45          & 92.93  & \textbf{97.47} & -      & -         & 97.01           &  \\
                       & F1      & 42.72 & 79.84          & 43.54 & 51.17   & 72.07          & 72.74  & 58.38          & -      & -         & \textbf{83.64}  &  \\
                       & p       & 45.45 & \textbf{85.65} & 46.37 & 56.19   & 77.50          & 75.90  & 64.25          & -      & -         & 83.34           &  \\
                       & r       & 47.76 & 80.60          & 50.75 & 53.03   & 72.73          & 73.74  & 62.63          & -      & -         & \textbf{83.58}  &  \\
                       & qwk     & 57.18 & 91.27          & 67.14 & 75.12   & 88.25          & 87.19  & 88.68          & -      & -         & \textbf{94.05}  &  \\ \cline{1-12}
\multirow{6}{*}{CMER}  & acc     & 23.06 & 28.38          & 22.62 & 22.20   & 26.40          & 26.11  & 28.61          & 26.50  & 26.50     & \textbf{48.89}  &  \\
                       & adj-acc & 50.78 & 59.87          & 49.67 & 46.46   & 57.37          & 53.10  & 56.64          & 58.50  & 62.47     & \textbf{76.99}  &  \\
                       & F1      & 20.28 & 27.28          & 17.51 & 19.64   & 26.61          & 24.87  & 26.39          & 25.16  & 22.06     & \textbf{48.51}  &  \\
                       & p       & 21.26 & 28.11          & 15.98 & 24.95   & 29.45          & 27.36  & 30.30          & -      & -         & \textbf{50.59}  &  \\
                       & r       & 23.06 & 28.38          & 22.62 & 22.20   & 26.40          & 26.11  & 28.61          & -      & -         & \textbf{48.89}  &  \\
                       & qwk     & 64.29 & 71.80          & 64.11 & 57.97   & 75.37          & 76.60  & 71.20          & -      & -         & \textbf{85.08}  &  \\ \cline{1-12}
\multirow{6}{*}{CLT}   & acc     & 30.40 & 42.00          & 35.20 & 29.33   & 41.47          & 41.73  & 37.87          & -      & -         & \textbf{46.00}  &  \\
                       & adj-acc & 76.80 & 81.60          & 78.80 & 64.80   & \textbf{85.60} & 84.00  & 75.73          & -      & -         & 84.80           &  \\
                       & F1      & 27.32 & 41.61          & 30.79 & 26.93   & 41.59          & 41.29  & 33.29          & -      & -         & \textbf{46.56}  &  \\
                       & p       & 27.92 & 42.71          & 34.42 & 28.53   & 44.53          & 46.02  & 34.88          & -      & -         & \textbf{50.19}  &  \\
                       & r       & 30.40 & 42.00          & 35.20 & 29.33   & 41.47          & 41.73  & 37.87          & -      & -         & \textbf{46.00}  &  \\
                       & qwk     & 74.98 & 85.17          & 77.69 & 54.85   & \textbf{85.22} & 83.40  & 73.59          & -      & -         & 84.93           &  \\ \cline{1-12}
\multirow{6}{*}{CTRDG} & acc     & 38.95 & 76.07          & 24.19 & 75.78   & 80.70          & 82.18  & 89.43          & -      & -         & \textbf{90.48}  &  \\
                       & adj-acc & 77.73 & 99.65          & 58.34 & 97.00   & 99.39          & 99.83  & \textbf{99.91} & -      & -         & 99.74           &  \\
                       & F1      & 27.94 & 75.96          & 10.43 & 76.04   & 80.66          & 82.18  & 89.18          & -      & -         & \textbf{90.50}  &  \\
                       & p       & 35.48 & 75.97          & 18.71 & 77.70   & 81.26          & 82.34  & 89.77          & -      & -         & \textbf{90.63}  &  \\
                       & r       & 38.95 & 76.07          & 24.19 & 75.78   & 80.70          & 82.18  & 89.43          & -      & -         & \textbf{90.48}  &  \\
                       & qwk     & 61.65 & 94.79          & 2.91  & 92.81   & 95.52          & 96.16  & 97.67          & -      & -         & \textbf{97.84}  &  \\ \cline{1-12}
\end{tabular}
}
\end{table*}

\section{Experimental Results \label{setc6}}

We report experimental results on all datasets (Section ~\ref{setc6.1}). We then present an ablation study (Section ~\ref{setc6.2}) and a comparison between single dimensional context weights (Section ~\ref{setc6.3}) and ordinal regression (Section ~\ref{setc6.4}) with DSDRRM.

\subsection{Overall Results \label{setc6.1}}

The experimental results of all models are summarized in Table~\ref{tab3}. Our DSDRRM achieves consistent improvements over the baselines on all datasets, which verifies the effectiveness of our proposed method. First, on the English dataset, our method has slightly lower p and qwk than the baseline model on the CEE dataset, but outperforms all baselines on other indicators on both English datasets, with OSP improving the accuracy by 2.8\% and CEE improving the accuracy by 2.98\%. To our surprise, on the Chinese dataset, our model improves the accuracy by 22.39\% over DTRA on the CMER dataset, and also improves to varying degrees on the CLT and CTRDG datasets. Second, in terms of the order of readability labels, our model achieves great improvements on qwk, which also verifies the effectiveness of the Ranking Model on ordered labels. Finally, surprisingly, RF seems to be more suitable for the evaluation of text readability, and the results on the Chinese dataset are even comparable to the neural network baseline, which illustrates the effectiveness of explicit features designed for Chinese datasets. This also lays the foundation for the future combination of explicit features and neural network features.

In the experiments, it can be observed that there are obvious differences in the accuracy of the DSDRRM model on four different readability datasets. The reason is that the structure of the dataset and the label standardization have an important impact on the model performance. There are differences in the modelability of the language itself. The grammatical rules of English are relatively fixed, and the subject-verb-object structure is clear. The granularity of Chinese language units is fuzzy, the syntax is flexible, and the semantic dependence is long, which makes modeling relatively difficult. In addition, the difference in the number of levels also significantly affects the performance of the model. As the number of classification levels increases, the model needs to make judgments in a finer-grained label space, and the classification difficulty increases accordingly. CMER has a total of 12 levels, which is far more than the 3 levels of OSP and the 6 levels of CTRDG. This increases the difficulty of learning the model, especially when the sample distribution is uneven. The model is more likely to be biased towards the prediction of the middle level, thereby reducing the overall accuracy.

\subsection{Ablation Study \label{setc6.2}}

In order to measure the contribution of multi-dimensional context weight, sentence tag assistance and ranking model to the model, we conducted ablation experiments on OSP and CLT. F1 and qwk were selected as evaluation indicators in the experiment. The experimental results are shown in Table~\ref{tab4}.

After removing the multi-dimensional context weight vector, the F1 and qwk of both datasets decreased, verifying that the multi-dimensional context weights contain useful information and can guide the attention model to locate information-rich words from the input sequence, thus playing an important role in the attention mechanism. After removing the sentence label auxiliary step, the performance of the model on the dataset also decreased. This result fully verifies the important role of sentence label assistance in improving document-level readability assessment. The strategy of sentence label-assisted document-level readability assessment provides an effective solution for fine-grained optimization of readability assessment tasks and significantly improves the generalization ability of the model. After removing the Ranking Model module, the OSP dataset has a significant decrease in F1 and qwk indicators. Due to the ordinal characteristics of readability labels, qwk also reflects the practicality of the Ranking Model module for ARA tasks.

\begin{table}[!t]
\caption{\label{tab4}Results of ablation experiments. -Context means removing multi-dimensional context weights. -MDEM means removing sentence tag assistance. -Ranking Model means removing the Ranking Model module.}
\centering
\begin{tabular}{cccccl}
\cline{1-5}
\multirow{2}{*}{Model} & \multicolumn{2}{c}{OSP} & \multicolumn{2}{c}{CLT} &  \\ \cline{2-5}
                       & F1         & qwk        & F1         & qwk        &  \\ \cline{1-5}
-Context               & 88.6       & 90.92      & 45.20      & 84.60      &  \\
-MDEM                  & 86.06      & 89.16      & 42.40      & 80.07      &  \\
-Ranking Model         & 87.58      & 90.68      & 45.69      & 84.40      &  \\
DSDRRM                 & 89.38      & 92.00      & 46.56      & 84.93      &  \\ \cline{1-5}
\end{tabular}
\end{table}

\subsection{Multi vs. Single Context: Comparative Analysis \label{setc6.3}}

In order to further verify the effectiveness of multidimensional context weights, we construct a single dimensional context weight for comparison experiments, that is, the same weight is used for each feature dimension in the word vector output by Bi-LSTM. Similar to Section 3.2.1, k convolution kernels are used to capture the output $h_i^t$ of Bi-LSTM to obtain the context information $\hat{c}=[c_1,\ldots,c_{m-l+1}]$. The pooling layer converts text of various lengths into fixed-length vectors. With the pooling layer, we can capture the information of the entire text. Therefore, the average pooling operation is applied to $\hat{c}$ to extract the average $a=mean(\hat{c})$. Finally, the sentence representation vector $h_i^s=a \cdot h_i^t$ is obtained by matrix multiplication of $a$ and $h_i^t$.

We conducted experiments using OSP and CLT, and the experimental results are shown in Table ~\ref{tab5}. Under multi-dimensional context weighting, although the model performance has a small loss in the qwk indicator on the OSP dataset, it has improved to varying degrees on the CLT dataset. In NLP, some tokens are polysemous. Since the traditional attention mechanism calculates the overall weight score of each word based on the word vector, it is impossible to distinguish the meaning of the same word in different contexts. The multi-dimensional weight vector calculates a weight score for each feature of each word, so it can select the feature that best describes the specific meaning of the word in any given context and include this information in the sentence encoding output.

\begin{table}[!t]
\caption{\label{tab5}Experimental results on Multi vs. Single Context. -SDW is represented with single dimensional context weights.}
\centering
\begin{tabular}{ccccc}
\hline
\multirow{2}{*}{Model} & \multicolumn{2}{c}{OSP} & \multicolumn{2}{c}{CLT} \\ \cline{2-5} 
                       & F1         & qwk        & F1         & qwk        \\ \hline
DSDRRM-SDW             & 89.33      & 92.20      & 45.16      & 83.88      \\
DSDRRM                 & 89.38      & 92.00      & 46.56      & 84.93      \\ \hline
\end{tabular}

\end{table}

\subsection{Ranking Model vs. Ordinal Regression \label{setc6.4}}

ARA can also be formulated as an ordinal regression task. Given a dataset with Y readability level categories, the document vector d represented by the model is input into a fully connected layer, outputting a readability label vector $r \in \mathbb{R}^Y$. The goal of ordinal regression is to minimize the ordinal regression loss, which is defined as follows:

\begin{equation}
    L(r;y)=-log(Sigmoid(\theta _k-r_k )-Sigmoid(\theta _{k-1}-r_{k-1} ))
\end{equation}
where $k=y$. $r_k$ denotes the $k^{th}$ dimension of the $r$. y is the true label. The threshold parameter $\theta_0,\ldots ,\theta_{Y-1}$ is also learned automatically from data. The probability of the current class is calculated by the threshold, and this probability is obtained by comparing the difference between the current class and the neighboring classes. This process is similar to the Ranking Model, understanding the ordinal relationship between categories by subtracting.

Similarly, we conduct experiments with OSP and CLT to further examine the effect of Ranking Model by comparing ordinal regression and multi-class classification. The experimental results are shown in Table~\ref{tab6}. No matter classification or ordinal regression, their results are worse than the Ranking model, which verifies the effectiveness of the pairwise comparison ranking algorithm. When comparing classification and ordinal regression, although ordinal regression is worse than classification on the weighted F1 measure, however, in terms of qwk, ordinal regression has a good improvement over classification on the whole. Therefore, it is also verified that the model can learn the differences between adjacent categories by doing subtraction, so as to better understand the sequential relationship between categories.

\begin{table}[!t]
\caption{\label{tab6}Model accuracy based on classification (C), ordinal regression (OR), and Ranking Model(RM).}
\centering
\begin{tabular}{ccccc}
\hline
\multirow{2}{*}{Model}      & \multicolumn{2}{c}{OSP}                               & \multicolumn{2}{c}{CLT}                               \\ \cline{2-5} 
                            & F1                        & qwk                       & F1                        & qwk                       \\ \hline
DSDR                        & 87.58                     & 90.68                     & 45.69                     & 84.40                     \\
\multicolumn{1}{l}{DSDR-OR} & \multicolumn{1}{l}{89.34} & \multicolumn{1}{l}{91.57} & \multicolumn{1}{l}{44.28} & \multicolumn{1}{l}{85.23} \\
DSDR-RM                     & 89.38                     & 92.00                     & 46.56                     & 84.93                     \\ \hline
\end{tabular}
\end{table}

\section{Conclusion and Future Work \label{setc7}}
This paper proposes a deep learning model for readability assessment of long documents. Compared to baseline models, our proposed forward and reverse readability assessment and pairwise sorting algorithms achieve competitive performance across all five datasets. In future work, we can further study the fusion of explicit  features and neural network features to improve the performance of readability assessment. At the same time, sentence labels from different corpora can be applied in cross-corpus evaluation, and more domain adaptation techniques can be considered to find the optimal feature set  capable of generalizing well to unseen texts.

\section*{Acknowledgments}
This work was supported by Guangdong Basic and Applied Basic Research Foundation [No. 022A1515011697]

\bibliographystyle{elsarticle-num-names}
\bibliography{refs}

@article{vogel1928objective,
  title={An objective method of determining grade placement of children's reading material},
  author={Vogel, Mabel and Washburne, Carleton},
  journal={The Elementary School Journal},
  volume={28},
  number={5},
  pages={373--381},
  year={1928},
  publisher={University of Chicago Press}
}

@article{cop2015eye,
  title={Eye movement patterns in natural reading: A comparison of monolingual and bilingual reading of a novel},
  author={Cop, Uschi and Drieghe, Denis and Duyck, Wouter},
  journal={PloS one},
  volume={10},
  number={8},
  pages={e0134008},
  year={2015},
  publisher={Public Library of Science San Francisco, CA USA},
  doi={10.1371/journal.pone.0134008}
}

@article{flesch1948new,
  title={A new readability yardstick.},
  author={Flesch, Rudolph},
  journal={Journal of applied psychology},
  volume={32},
  number={3},
  pages={221},
  year={1948},
  publisher={American Psychological Association}
}

@article{dale1948formula,
  title={A formula for predicting readability: Instructions},
  author={Dale, Edgar and Chall, Jeanne S},
  journal={Educational research bulletin},
  pages={37--54},
  year={1948},
  publisher={JSTOR}
}

@article{mc1969smog,
  title={SMOG grading-a new readability formula},
  author={Mc Laughlin, G Harry},
  journal={Journal of reading},
  volume={12},
  number={8},
  pages={639--646},
  year={1969},
  publisher={JSTOR}
}

@inproceedings{villata2020plain,
  title={Plain language assessment of statutes},
  author={Villata, S and others},
  booktitle={Legal Knowledge and Information Systems: JURIX 2020: The Thirty-third Annual Conference, Brno, Czech Republic, December 9-11, 2020},
  volume={334},
  pages={207},
  year={2020},
  organization={IOS Press}
}

@article{sare2020readability,
  title={Readability assessment of Internet-based patient education materials related to treatment options for benign prostatic hyperplasia},
  author={Sare, Antony and Patel, Aesha and Kothari, Pankti and Kumar, Abhishek and Patel, Nitin and Shukla, Pratik A},
  journal={Academic Radiology},
  volume={27},
  number={11},
  pages={1549--1554},
  year={2020},
  publisher={Elsevier},
  doi = {10.1002/lary.23424}

}

@article{perni2019assessment,
  title={Assessment of use, specificity, and readability of written clinical informed consent forms for patients with cancer undergoing radiotherapy},
  author={Perni, Subha and Rooney, Michael K and Horowitz, David P and Golden, Daniel W and McCall, Anne R and Einstein, Andrew J and Jagsi, Reshma},
  journal={JAMA oncology},
  volume={5},
  number={8},
  pages={e190260--e190260},
  year={2019},
  publisher={American Medical Association},
doi={10.1001/jamaoncol.2019.0260}
}

@article{vajjala2021trends,
  title={Trends, limitations and open challenges in automatic readability assessment research},
  author={Vajjala, Sowmya},
  journal={arXiv preprint arXiv:2105.00973},
  year={2021},
doi={10.48550/arxiv.2105.00973 }
}

@inproceedings{schwarm2005reading,
  title={Reading level assessment using support vector machines and statistical language models},
  author={Schwarm, Sarah E and Ostendorf, Mari},
  booktitle={Proceedings of the 43rd annual meeting of the Association for Computational Linguistics (ACL’05)},
  pages={523--530},
  year={2005},
 doi={10.3115/1219840.1219905 }
}

@inproceedings{lee2021pushing,
    title = "Pushing on Text Readability Assessment: A Transformer Meets Handcrafted Linguistic Features",
    author = "Lee, Bruce W.  and
      Jang, Yoo Sung  and
      Lee, Jason",
    editor = "Moens, Marie-Francine  and
      Huang, Xuanjing  and
      Specia, Lucia  and
      Yih, Scott Wen-tau",
    booktitle = "Proceedings of the 2021 Conference on Empirical Methods in Natural Language Processing",
    month = nov,
    year = "2021",
    address = "Online and Punta Cana, Dominican Republic",
    publisher = "Association for Computational Linguistics",
    url = "https://aclanthology.org/2021.emnlp-main.834/",
    doi = "10.18653/v1/2021.emnlp-main.834",
    pages = "10669--10686",
}

@inproceedings{hansen2021machine,
  title={Machine Learning for Readability Assessment and Text Simplification in Crisis Communication: A Systematic Review},
  author={Hansen, Hieronymus and Widera, Adam and Ponge, Johannes and Hellingrath, Bernd},
  booktitle={Proceedings of the 54th Hawaii International Conference on System Sciences},
  year={2021},
  doi={10.24251/hicss.2021.277}
}

@inproceedings{pilan2016predicting,
  title={Predicting proficiency levels in learner writings by transferring a linguistic complexity model from expert-written coursebooks},
  author={Pil{\'a}n, Ildik{\'o} and Volodina, Elena and Zesch, Torsten},
  booktitle={Proceedings of COLING 2016, the 26th International Conference on Computational Linguistics: Technical Papers},
  pages={2101--2111},
  year={2016}
}

@article{li2022unified,
  title={A Unified Neural Network Model for Readability Assessment with Feature Projection and Length - Balanced Loss},
  author={Li, Wen Biao and Wang, Ziyang and Wu, Yunfang},
  journal={arXiv preprint arXiv:2210.10305},
  year={2023},
  doi={10.48550/arxiv.2210.10305}
}

@article{martinc2021supervised,
  title={Supervised and unsupervised neural approaches to text readability},
  author={Martinc, Matej and Pollak, Senja and Robnik-{\v{S}}ikonja, Marko},
  journal={Computational Linguistics},
  volume={47},
  number={1},
  pages={141--179},
  year={2021},
  publisher={MIT Press},
  doi={10.1162/COLI_A_00398 }
}

@inproceedings{imperial2021bert,
  title={BERT Embeddings for Automatic Readability Assessment},
  author={Imperial, Joseph Marvin},
  booktitle={Proceedings of the International Conference on Recent Advances in Natural Language Processing},
  pages={611--618},
  year={2021},
  doi={10.26615/978 - 954 - 452 - 072 - 4_069}
}

@article{lee2023prompt,
  title={Prompt-based learning for text readability assessment},
  author={Lee, Bruce W and Lee, Jason Hyung-Jong},
  journal={arXiv preprint arXiv:2302.13139},
  year={2023},
  doi={10.48550/arxiv.2302.13139 }
}

@inproceedings{risch2020bagging,
  title={Bagging BERT Models for Robust Aggression Identification},
  author={Risch, Julian and Krestel, Ralf},
  booktitle={Proceedings of the Second Workshop on Trolling, Aggression and Cyberbullying},
  pages={55--61},
  year={2020},
  address={Marseille, France},
  publisher={European Language Resources Association (ELRA)},
  doi={10.5281/zenodo.3727018},
  url={https://aclanthology.org/2020.trac-1.9/}
}

@inproceedings{kenton2019bert,
  title={BERT: Pre-training of Deep Bidirectional Transformers for Language Understanding},
  author={Kenton, Jacob Devlin Ming-Wei Chang and Toutanova, Lee Kristina},
  booktitle={Proceedings of NAACL-HLT},
  pages={4171--4186},
  year={2019}, 
doi={10.18653/v1/N19-1423 }
}

@inproceedings{yang2016hierarchical,
  author = {Yang, Zichao and Yang, Diyi and Dyer, Chris and He, Xiaodong and Smola, Alex and Hovy, Eduard},
  title = {Hierarchical Attention Networks for Document Classification},
  booktitle = {Proceedings of the 2016 Conference of the North American Chapter of the Association for Computational Linguistics: Human Language Technologies},
  year = {2016},
  publisher = {Association for Computational Linguistics},
  pages = {1480--1489},
  location = {San Diego, California},
  doi = {10.18653/v1/N16-1174},
  url = {http://www.aclweb.org/anthology/N16-1174}
}

@article{zheng2019hybrid,
  title={A hybrid bidirectional recurrent convolutional neural network attention - based model for text classification},
  author={Zheng, Jin and Zheng, Limin},
  journal={IEEE Access},
  year={2019},
  volume={7},
  pages={106673--106685},
  doi={10.1109/access.2019.2932619}
}

@article{azpiazu2019multiattentive,
  title={Multiattentive recurrent neural network architecture for multilingual readability assessment},
  author={Azpiazu, Ion Madrazo and Pera, Maria Soledad},
  journal={Transactions of the Association for Computational Linguistics},
  volume={7},
  pages={421--436},
  year={2019},
  publisher={MIT Press},
  doi={10.1162/tacl_a_00278},
  url={https://aclanthology.org/Q19-1028/}
}

@inproceedings{lee2022neural,
  author = {Lee, Justin and Vajjala, Sowmya},
  title = {A Neural Pairwise Ranking Model for Readability Assessment},
  booktitle = {Findings of the Association for Computational Linguistics: ACL 2022},
  year = {2022},
  publisher = {Association for Computational Linguistics},
  pages = {3802--3813},
  location = {Dublin, Ireland},
  doi = {10.18653/v1/2022.findings-acl.300}
}

@article{kincaid1975derivation,
  title={Derivation of new readability formulas (automated readability index, fog count and flesch reading ease formula) for navy enlisted personnel},
  author={Kincaid, J Peter and Fishburne Jr, Robert P and Rogers, Richard L and Chissom, Brad S},
  year={1975},
  publisher={Institute for Simulation and Training, University of Central Florida}
}

@inproceedings{heilman2008analysis,
  title={An analysis of statistical models and features for reading difficulty prediction},
  author={Heilman, Michael and Collins-Thompson, Kevyn and Eskenazi, Maxine},
  booktitle={Proceedings of the third workshop on innovative use of NLP for building educational applications},
  pages={71--79},
  year={2008}
}

@inproceedings{feng2010comparison,
  title={A comparison of features for automatic readability assessment},
  author={Feng, Lijun and Jansche, Martin and Huenerfauth, Matt and Elhadad, No{\'e}mie},
  booktitle={Coling 2010: Posters},
  pages={276--284},
  year={2010}
}

@inproceedings{hancke2012readability,
  title={Readability classification for German using lexical, syntactic, and morphological features},
  author={Hancke, Julia and Vajjala, Sowmya and Meurers, Detmar},
  booktitle={Proceedings of COLING 2012},
  pages={1063--1080},
  year={2012}
}

@inproceedings{qiu2018exploring,
  title={Exploring the impact of linguistic features for Chinese readability assessment},
  author={Qiu, Xinying and Deng, Kebin and Qiu, Likun and Wang, Xin},
  booktitle={Natural Language Processing and Chinese Computing: 6th CCF International Conference, NLPCC 2017, Dalian, China, November 8--12, 2017, Proceedings 6},
  pages={771--783},
  year={2018},
  organization={Springer},
  doi={10.1007/978-3-319-73618-1_67}
}

@article{deutsch2020linguistic,
  title={Linguistic features for readability assessment},
  author={Deutsch, Tovly and Jasbi, Masoud and Shieber, Stuart},
  journal={arXiv preprint arXiv:2006.00377},
  year={2020},
  DOI={10.18653/v1/2020.bea-1.1}
}

@inproceedings{jiang2015graph,
  title={A graph-based readability assessment method using word coupling},
  author={Jiang, Zhiwei and Sun, Gang and Gu, Qing and Bai, Tao and Chen, Daoxu},
  booktitle={Proceedings of the 2015 Conference on Empirical Methods in Natural Language Processing},
  pages={411--420},
  year={2015},
  doi={10.18653/v1/D15-1047}
}

@article{blaneck2022automatic,
  title={Automatic readability assessment of German sentences with transformer ensembles},
  author={Blaneck, Patrick Gustav and Bornheim, Tobias and Grieger, Niklas and Bialonski, Stephan},
  journal={arXiv preprint arXiv:2209.04299},
  year={2022}
}

@inproceedings{meng2020readnet,
  title={Readnet: A hierarchical transformer framework for web article readability analysis},
  author={Meng, Changping and Chen, Muhao and Mao, Jie and Neville, Jennifer},
  booktitle={Advances in Information Retrieval: 42nd European Conference on IR Research, ECIR 2020, Lisbon, Portugal, April 14--17, 2020, Proceedings, Part I 42},
  pages={33--49},
  year={2020},
  organization={Springer},
  doi={10.1007/978-3-030-45439-5_3}
}

@inproceedings{rennie2005loss,
  title={Loss functions for preference levels: Regression with discrete ordered labels},
  author={Rennie, Jason DM and Srebro, Nathan},
  booktitle={Proceedings of the IJCAI multidisciplinary workshop on advances in preference handling},
  volume={1},
  year={2005},
  organization={AAAI Press, Menlo Park, CA}
}

@inproceedings{zeng2022enhancing,
  title={Enhancing automatic readability assessment with pre-training and soft labels for ordinal regression},
  author={Zeng, Jinshan and Xie, Yudong and Yu, Xianglong and Lee, John SY and Zhou, Ding-Xuan},
  booktitle={Findings of the Association for Computational Linguistics: EMNLP 2022},
  pages={4557--4568},
  year={2022},
  doi={10.18653/v1/2022.findings-emnlp.334}
}

@inproceedings{diaz2019soft,
  title={Soft labels for ordinal regression},
  author={Diaz, Raul and Marathe, Amit},
  booktitle={Proceedings of the IEEE/CVF conference on computer vision and pattern recognition},
  pages={4738--4747},
  year={2019},
  doi={10.1109/CVPR.2019.00487}
}

@article{tanaka2010sorting,
  title={Sorting texts by readability},
  author={Tanaka-Ishii, Kumiko and Tezuka, Satoshi and Terada, Hiroshi},
  journal={Computational linguistics},
  volume={36},
  number={2},
  pages={203--227},
  year={2010},
  publisher={MIT Press One Rogers Street, Cambridge, MA 02142-1209, USA journals-info~…},
  doi={10.1162/coli.09-036-R2-08-050}
}

@article{xia2019text,
  title={Text readability assessment for second language learners},
  author={Xia, Menglin and Kochmar, Ekaterina and Briscoe, Ted},
  journal={arXiv preprint arXiv:1906.07580},
  year={2019},
  doi={
https://doi.org/10.18653/v1/W16-0502}
}

@article{madrazo2020cross,
  title={Is cross-lingual readability assessment possible?},
  author={Madrazo Azpiazu, Ion and Pera, Maria Soledad},
  journal={Journal of the Association for Information Science and Technology},
  volume={71},
  number={6},
  pages={644--656},
  year={2020},
  publisher={Wiley Online Library},
  doi={10.1002/asi.24293}
}

@article{sung2013investigating,
  title={Investigating Chinese text readability: linguistic features, modeling, and validation.},
  author={Sung, Yao-Ting and Chen, Ju-Ling and Lee, Yi-Shian and Cha, Jih-Ho and Tseng, Hou-Chiang and Lin, Wei-Chun and Chang, Tao-Hsing and Chang, Kuo-En},
  journal={Chinese Journal of Psychology},
  year={2013},
  publisher={Taiwanese Psychological Assn}
}

@inproceedings{ma2022research,
  title={Research on the Evaluation of the Classical Chinese Difficulty in the Compulsory Education Stage},
  author={Ma, Kun and Liu, Zhiying and Yang, Lijiao and Sun, Ning and Wang, Yujun and Qiu, Ziliang},
  booktitle={2022 International Conference on Asian Language Processing (IALP)},
  pages={353--357},
  year={2022},
  organization={IEEE}
}

@article{du2019convolution,
  title={Convolution-based neural attention with applications to sentiment classification},
  author={Du, Jiachen and Gui, Lin and He, Yulan and Xu, Ruifeng and Wang, Xuan},
  journal={IEEE Access},
  volume={7},
  pages={27983--27992},
  year={2019},
  publisher={IEEE},
  doi={10.1109/ACCESS.2019.2900335}
}

@article{vaswani2017attention,
  title={Attention is all you need},
  author={Vaswani, Ashish and Shazeer, Noam and Parmar, Niki and Uszkoreit, Jakob and Jones, Llion and Gomez, Aidan N and Kaiser, {\L}ukasz and Polosukhin, Illia},
  journal={Advances in neural information processing systems},
  volume={30},
  year={2017}
}

@inproceedings{hu2021hierarchical,
  title={Hierarchical attention transformer networks for long document classification},
  author={Hu, Yongli and Chen, Puman and Liu, Tengfei and Gao, Junbin and Sun, Yanfeng and Yin, Baocai},
  booktitle={2021 International Joint Conference on Neural Networks (IJCNN)},
  pages={1--7},
  year={2021},
  organization={IEEE},
  doi={10.1109/IJCNN52387.2021.9533869}
}

@inproceedings{shen2018disan,
  title={Disan: Directional self-attention network for rnn/cnn-free language understanding},
  author={Shen, Tao and Zhou, Tianyi and Long, Guodong and Jiang, Jing and Pan, Shirui and Zhang, Chengqi},
  booktitle={Proceedings of the AAAI conference on artificial intelligence},
  volume={32},
  number={1},
  year={2018},
  doi={10.48550/arXiv.1709.04696}
}

@article{xie2020unsupervised,
  title={Unsupervised data augmentation for consistency training},
  author={Xie, Qizhe and Dai, Zihang and Hovy, Eduard and Luong, Thang and Le, Quoc},
  journal={Advances in neural information processing systems},
  volume={33},
  pages={6256--6268},
  year={2020}
}

@mastersthesis{ 1023569814.nh,
author = { Wenbiao Li },
title = {Research on text readability evaluation based on neural network model},
school = {Peking University},
year = {2023}
}

@inproceedings{vajjala2018onestopenglish,
  title={OneStopEnglish corpus: A new corpus for automatic readability assessment and text simplification},
  author={Vajjala, Sowmya and Lu{\v{c}}i{\'c}, Ivana},
  booktitle={Proceedings of the thirteenth workshop on innovative use of NLP for building educational applications},
  pages={297--304},
  year={2018},
  doi={10.18653/v1/W18-0535}
}

@article{tan2024multi,
  author  = {Tan, Keren and Lan, Yunshi and Zhang, Yang and Ding, Anqi},
  title   = {A Chinese Text Readability Classification Model Based on Multi-Level Linguistic Feature Fusion},
  journal = {Journal of Chinese Information Processing},
  volume  = {38},
  number  = {05},
  pages   = {41--52},
  year    = {2024},
  issn    = {1003-0077}
}

@article{pedregosa2011scikit,
  title={Scikit-learn: Machine learning in Python},
  author={Pedregosa, Fabian and Varoquaux, Ga{\"e}l and Gramfort, Alexandre and Michel, Vincent and Thirion, Bertrand and Grisel, Olivier and Blondel, Mathieu and Prettenhofer, Peter and Weiss, Ron and Dubourg, Vincent and others},
  journal={the Journal of machine Learning research},
  volume={12},
  pages={2825--2830},
  year={2011},
  publisher={JMLR. org},
  doi={10.5555/1953048.2078195}
}

@article{lan2019albert,
  title={Albert: A lite bert for self-supervised learning of language representations},
  author={Lan, Zhenzhong and Chen, Mingda and Goodman, Sebastian and Gimpel, Kevin and Sharma, Piyush and Soricut, Radu},
  journal={arXiv preprint arXiv:1909.11942},
  year={2019}
}

@article{paszke2019pytorch,
  title={Pytorch: An imperative style, high-performance deep learning library},
  author={Paszke, Adam and Gross, Sam and Massa, Francisco and Lerer, Adam and Bradbury, James and Chanan, Gregory and Killeen, Trevor and Lin, Zeming and Gimelshein, Natalia and Antiga, Luca and others},
  journal={Advances in neural information processing systems},
  volume={32},
  year={2019}
}

@inproceedings{ehara2021evaluation,
  title={Evaluation of unsupervised automatic readability assessors using rank correlations},
  author={Ehara, Yo},
  booktitle={Proceedings of the 2nd Workshop on Evaluation and Comparison of NLP Systems},
  pages={62--72},
  year={2021},
  doi={10.18653/v1/2021.eval4nlp-1.7}
}

@article{kingma2014adam,
  title={Adam: A method for stochastic optimization},
  author={Kingma, Diederik P and Ba, Jimmy},
  journal={arXiv preprint arXiv:1412.6980},
  year={2014},
  doi={10.48550/arXiv.1412.6980}
}

\appendix

\section{Chinese Explicit features Details}

\setcounter{table}{0}  

\begin{table}[!h]
\caption{\label{tab8}Lexical class features}
\centering
\begin{tabular}{cc}
\hline
\textbf{Categories}                       & \textbf{Feature name}                          \\ \hline
\multirow{2}{*}{Number of words}          & Word count                                     \\
                                          & Word count                                     \\ \hline
\multirow{2}{*}{Vocabulary richness}      & Dissimilar Word Ratio (TTR)                    \\
                                          & Content word density                           \\ \hline
\multirow{2}{*}{Word frequency}           & Log average of real word frequencies           \\
                                          & Number of difficult words                      \\ \hline
\multirow{6}{*}{Vocabulary length}        & Low stroke number of characters                \\
                                          & The number of characters in   the stroke       \\
                                          & Number of high stroke   characters             \\
                                          & Average number of strokes per   character      \\
                                          & Number of two-character words                  \\
                                          & More than three words                          \\ \hline
\multirow{2}{*}{Semantic class   metrics} & Number of content words                        \\
                                          & Negative words                                 \\ \hline
\multirow{4}{*}{Syntactic metrics}        & Average number of words in simple sentences    \\
                                          & Average number of words in   complex sentences \\
                                          & Single sentence ratio                          \\
                                          & Noun phrase ratio                              \\ \hline
\multirow{2}{*}{Vocabulary density}       & Number of function words                       \\
                                          & Function word density                          \\ \hline
\multirow{2}{*}{Diversity of words}       & RTTR                                           \\
                                          & MTLD                                           \\ \hline
\end{tabular}
\end{table}

\begin{table}[!h]
\caption{\label{tab9}Part of speech feature}
\centering
\begin{tabular}{cc}
\hline
\textbf{Categories}        & \textbf{Feature name}                              \\ \hline
\multirow{5}{*}{Adjective} & Percentage of adjectives                           \\
                           & Percentage of unique   adjectives                  \\
                           & Number of unique adjectives                        \\
                           & Average number of adjectives   per sentence        \\
                           & Average number of unique   adjectives per sentence \\ \hline
\multirow{5}{*}{Noun}      & Percentage of nouns                                \\
                           & Percentage of unique nouns                         \\
                           & Number of unique nouns                             \\
                           & Average number of nouns per   sentence             \\
                           & Average number of unique nouns   per sentence      \\ \hline
\multirow{5}{*}{Verb}      & Percentage of verbs                                \\
                           & Percentage of unique verbs                         \\
                           & Number of unique verbs                             \\
                           & Average number of verbs per   sentence             \\
                           & Average number of unique verbs   per sentence      \\ \hline
\end{tabular}
\end{table}

\begin{table}[!h]
\caption{\label{tab10}Discourse feature}
\centering
\begin{tabular}{cc}
\hline
\textbf{Categories}            & \textbf{Feature name}                           \\ \hline
\multirow{6}{*}{Solid density} & Number of named entities                        \\
                               & Number of unique named   entities               \\
                               & Percentage of named entities                    \\
                               & Percentage of unique named   entities           \\
                               & Average number of named   entities per sentence \\
                               & Number of unique named   entities per sentence  \\ \hline
\end{tabular}
\end{table}

\begin{table}[!h]
\caption{\label{tab11}Article cohesion characteristics}
\centering
\begin{tabular}{cc}
\hline
\textbf{Categories}               & \textbf{Feature name} \\ \hline
Term of reference                 & Number of pronouns    \\
\multirow{3}{*}{Connective words} & Number of connectives \\
                                  & Positive connectives  \\
                                  & Negative connectives  \\ \hline
\end{tabular}
\end{table}

\end{document}